\pdfoutput=1

\documentclass[11pt]{article}

\usepackage{acl}

\usepackage{times}
\usepackage{latexsym}

\usepackage[T1]{fontenc}

\usepackage[utf8]{inputenc}

\usepackage{microtype}

\usepackage{inconsolata}

\usepackage{graphicx}
\usepackage{enumerate}
\usepackage[colorinlistoftodos]{todonotes}
\usepackage{booktabs}
\usepackage{multirow}
\usepackage{hyperref}
\usepackage{caption}
\usepackage{subcaption}
\usepackage{amssymb}
\usepackage{array}
\usepackage{nicematrix}
\usepackage{comment}
\usepackage{nicematrix}

\title{The Road to Quality is Paved with Good Revisions: A Detailed Evaluation Methodology for Revision Policies in Incremental Sequence Labelling}

\author{Brielen Madureira$^{\mathbf{1}}$ \hspace{10mm}  Patrick Kahardipraja$^{\mathbf{1}}$  \hspace{10mm}  David Schlangen$^{\mathbf{1, 2}}$ \\
$^{\mathbf{1}}$Computational Linguistics, Department of Linguistics, University of Potsdam, Germany \\
$^{\mathbf{2}}$German Research Center for Artificial Intelligence (DFKI), Berlin, Germany \\
  \texttt{\{madureiralasota,kahardipraja,david.schlangen\}@uni-potsdam.de}}

\begin{document}
\maketitle
\begin{abstract}
Incremental dialogue model components produce a sequence of output prefixes based on incoming input. Mistakes can occur due to local ambiguities or to wrong hypotheses, making the ability to revise past outputs a desirable property that can be governed by a policy. In this work, we formalise and characterise edits and revisions in incremental sequence labelling and propose metrics to evaluate revision policies. We then apply our methodology to profile the incremental behaviour of three Transformer-based encoders in various tasks, paving the road for better revision policies.
\end{abstract}

\section{Introduction}
\label{sec:intro}

Since the dawn of Wikipedia, users have made $1.7 \times 10^9$ edits to its pages. Its most revised entry contains 56,713 revisions, all documented in the page history.\footnote{According to \href{https://stats.wikimedia.org/\#/all-wikipedia-projects/contributing/user-edits/normal|table|2001-01-01~2023-05-01|(page\_type)~content*non-content|monthly}{Wikimedia Statistics} and \href{https://en.wikipedia.org/wiki/Special:MostRevisions}{wiki Special}.} In such an active community, conflicts inevitably occur. Editors can begin competing to override each other's contributions, causing dysfunctional \textit{edit warrings}.\footnote{\href{https://en.wikipedia.org/wiki/Wikipedia:Edit\_warring}{https://en.wikipedia.org/wiki/Wikipedia:Edit\_warring}} To help regulate the environment, an editing policy is in force, aiming at making edits constructive and improving quality.\footnote{\href{https://en.wikipedia.org/wiki/Wikipedia:Editing_policy}{https://en.wikipedia.org/wiki/Wikipedia:Editing\_policy}}

Edits, revisions and policies are key concepts in incremental processing, where a model must rely on partial input to generate partial output. Incrementality can help optimise reactivity, naturalness, quality and realism in interactive settings \citep{iu-restart}. This is particularly relevant in dialogue models whose NLU components need to operate on incoming input, \textit{e.g.}~while performing NER, slot filling or disfluency detection, or doing simultaneous translation. 

Local ambiguities in the linguistic input and transient mistakes by the model can result in wrong partial hypotheses, so that the ability to \textit{revise}, by \textit{editing} previous outputs, is desirable \citep{tapir}. Beyond monitoring the occurrence of edits, it is also beneficial to have a \textit{policy} regulating when and which revisions should be made, reducing the occurrence of undesirable edits. Existing literature using consolidated incremental evaluation metrics falls short in capturing relevant nuances of the incremental behaviour in terms of revisions.  

\begin{figure}[t]
    \centering
    \includegraphics[trim={0cm 2cm 8.9cm 0cm},clip,width=0.78\columnwidth,page=1]{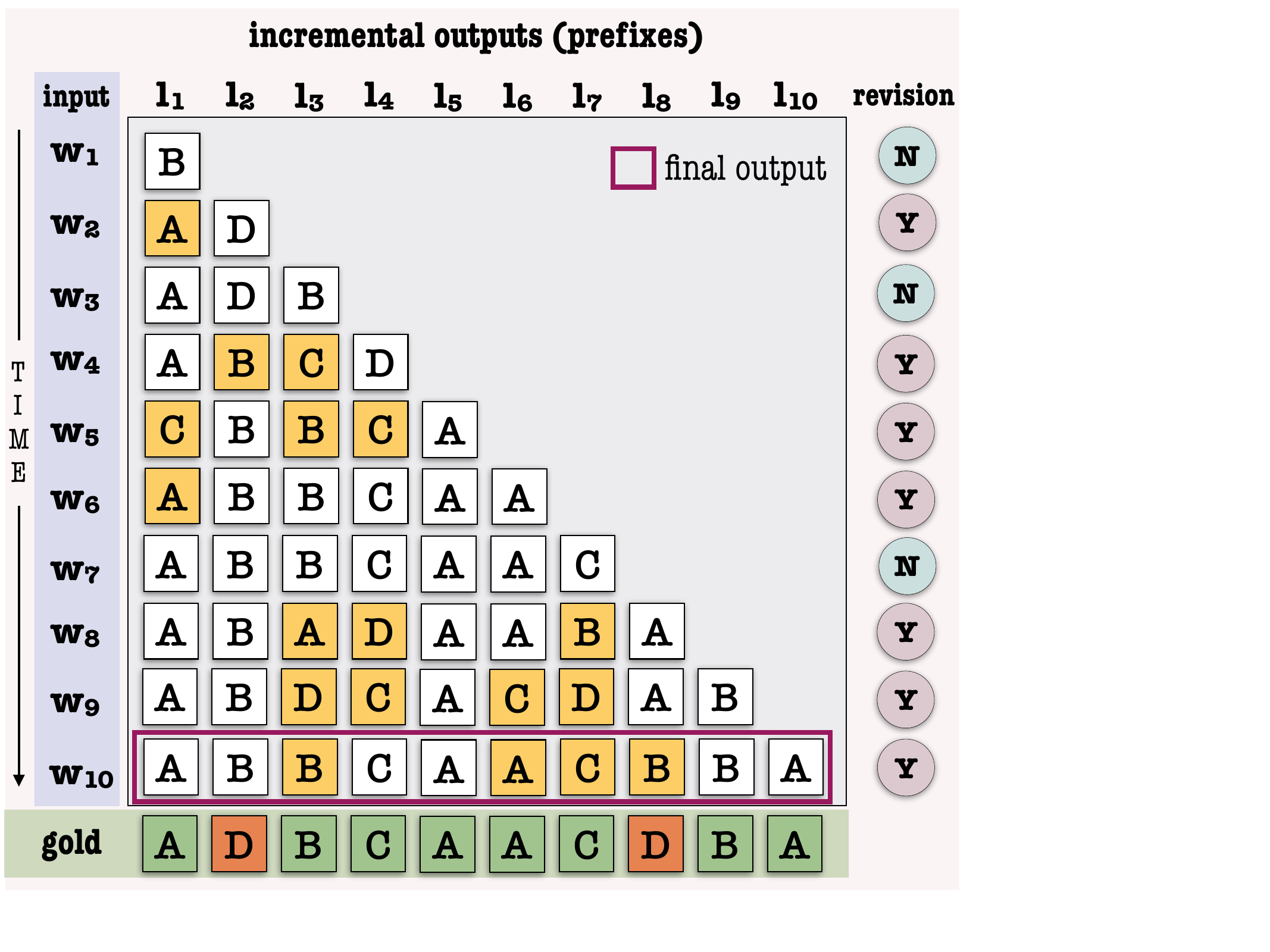}
    \caption{Constructed example of an incremental chart containing output prefixes with marked edits (yellow) and revisions in incremental sequence labelling. Red stands for wrong final predictions wrt. the gold standard.}
    \label{fig:contructed-example}
\end{figure}

In this work, we propose an evaluation methodology for revision policies in incremental sequence labelling. A constructed example is shown in Figure~\ref{fig:contructed-example}, with revisions indicated in the right column. Specifically, our contributions to address the identified evaluation gap are: A formalisation of revision policy in incremental sequence labelling, characterising types of edits and of revisions (\S \ref{sec:formalisation}-\ref{sec:characterisation}); a proposal of specialised evaluation metrics for revision policies, accompanied by a discussion on the desired behaviour of incremental processors (\S\ref{sec:metrics}-\ref{sec:ideal}); and a demonstration of our methodology with an analysis of the revision policy in three sequence labelling Transformer-based models (\S \ref{sec:analysis}).\footnote{Our implementation is available at \url{https://github.com/briemadu/inc-eval-revisions} with accompanying documentation on how to run the evaluation for other models.}

\section{Motivation}
\label{sec:motivation}
Incremental natural language processing\footnote{For a review, see \citet{kohn-2018-incremental}. In other contexts, also referred to as real-time processing \citep{pozzan2015revise} or streaming \citep{kaushal-etal-2023-efficient}.} has \textit{time} at front line, being pivotal for interactive settings. At each time step, models must operate on partial input to deliver partial output, but sometimes previous decisions have to be revised. For example, at time step $4$ in Figure \ref{fig:contructed-example}, the labels for the input tokens $2$ and $3$ were edited into new states. With regard to revisions, at least three types of incremental processors exist, as summarised in Table \ref{table:types}:

\begin{enumerate}
    \item Inherently incremental but monotonic models. They keep an internal state that is updated and used to extend the output at each time step, but cannot revise previous outputs. 

    \item Non-incremental models used with a \textit{restart-incremental} interface, being forced to perform a full recomputation at each time step. Such models revise the output as a by-product of their recomputations.

    \item Incremental models with a dedicated policy to detect the need to perform revisions only when deemed necessary and, more specifically, deciding which parts of the output prefix need to be revised and how.
\end{enumerate}

\begin{table}[h!]
    \centering
    \small
    \begin{tabular}{@{}m{0.05cm} m{0.2cm} m{2.8cm} m{2.5cm}@{}} 
    \toprule
     &  & \textbf{non-incremental} &  \textbf{incremental}   \\
     \cmidrule{3-4} 
     \multirow{2}{*}{\rotatebox[origin=c]{90}{\textbf{revisions}}}    & no & n/a &  strictly monotonic outputs \\
     \cmidrule{3-4} 
                                   &  yes  & recomputation policy doing revisions as a by-product  & revision policy     \\              
    \bottomrule
    \end{tabular}
    \caption{Types of incremental processors.}
    \label{table:types}
\end{table}

Monotonicity avoids instability in the output, allowing subprocesses to start immediately, as it is certain that the outputs will not change. However, they never recover from mistakes, which is one of the drawbacks of employing vanilla RNNs and LSTMs \citep{lstm}.

Models that depend on the availability of full sentences at once can be ``incrementalised'' with the \textit{restart-incremental} paradigm \citep{iu-restart}, causing revisions to occur via recomputations.\footnote{Also called \textit{incremental interface} \citep{beuck-etal-2011-decision} or \textit{beat-driven approach} 
\citep{baumann-incremental}.} 

Cutting-edge NLP models currently rely on Transformers \citep{vaswani2017attention}, which are non-incremental. Using them in a \textit{restart-incremental} fashion requires recomputing from scratch at every time step, which we hereby name the \textit{naive recomputation policy}. It is a very expensive policy because, for a sequence of $n$ tokens, the complexity is $\sum_{i=1}^{n}i^2$ (\textit{i.e.}~the $n$-th square pyramidal number). Besides, this naive approach wastes computational budget, because not all recomputations cause revisions. The results reported by \citet{tapir}, for example, show that only around 25\% of the recomputations actually changed the output prefix. The disadvantages of the naive policy can be alleviated by a smarter policy that cuts down the number of time steps with recomputations.   

Still, beyond deciding when to \textit{recompute}, a revision policy par excellence should directly guide the more specific decision of when (and what) to actually \textit{revise}, and must be evaluated accordingly.

\section{Related Literature}
\label{sec:litreview}
\begin{table*}[ht!]
\centering
\footnotesize

\begin{NiceTabular}{m{7cm} m{4cm} m{3.5cm}}
\CodeBefore
\rowcolors{2}{white}{gray!25}
\Body
    \toprule
        latency, quality, stability & simultaneous translation &  \citet{arivazhagan-etal-2020-translation} \citet{ma-etal-2020-simuleval}
        \\ \midrule    
        quality, responsiveness, robustness, stability & speech recognition and \newline  diarization & \citet{baumann-etal-2009-assessing} \newline \citet{addlesee-etal-2020-comprehensive} \\
        \midrule
        similarity, timing, diachronic & general & \citet{baumann-incremental} \\ 
        \midrule
        fluency, latency, quality, recovery capabilities, timing & simultaneous interpreting \newline (MT and speech synthesis) & \citet{baumann-etal-2014-towards} \\ 
        \midrule
        decisiveness, monotonicity, stability, timeliness & POS tagging & \citet{beuck-etal-2011-decision} \\
        \midrule
        amount of predicted information, connectedness, delay, inclusiveness, monotonicity, quality & parsing & \citet{beuck-inc-parsing, beuck2013predictive} \citet{kohn-menzel-2014-incremental} \\
        \midrule
        cognitive aspects, efficiency &  
        neural coreference resolution & \citet{grenander-etal-2022-sentence} \\
        \midrule
        jumpiness, position & reference resolution & \citet{schlangen-etal-2009-incremental} \\
        \midrule 
        accuracy, integration, representational similarity & sequence-to-sequence & \citet{ulmer-etal-2019-assessing}  \\
        \midrule
        consistency, diminishing returns, interruptibility, monotonicity, preemptability, (recognisable) quality & anytime algorithms & \citet{zilberstein1996using} \\
    \bottomrule

    \end{NiceTabular}%
    \caption{Overview of relevant properties for incremental evaluation in various tasks.}
    \label{table:properties}
\end{table*}

Revisability is in the nature of incremental processing: Hypothesis revision is a necessary operation to correct mistakes and build up a high-quality final output \citep{iu-restart}. Still, there is a trade-off between requiring that later modules handle a processor's revisions and buying stability by reducing some of its incrementality, which makes the concept of \textit{hypothesis stability} very relevant \citep{baumann-etal-2009-assessing}. \citet{beuck-etal-2011-decision} argue that performing revisions should not take as long as the initial processing, so as to retain the advantages of incremental processing. They propose two strategies: Allowing revisions only within a fixed window or limiting their types. Empirically determining how often a model changes the output is an aspect of their analysis we also rely on.

 The restart-incremental paradigm was investigated for Transformer-based sequence labelling by \citet{madureira-schlangen-2020-incremental} and \citet{kahardipraja-etal-2021-towards}; recently, adaptive policies were proposed to reduce the computational load \citep{kaushal-etal-2023-efficient,tapir}. \citet{rohanian-hough-2021-best} and \citet{chen-etal-2022-teaching} explored adaptation strategies to use Transformers for incremental disfluency detection. In simultaneous translation, where policies are a central concept \citep{zheng-etal-2020-simultaneous,zhang-etal-2020-learning-adaptive}, the restart-incremental approach is in use and revisions are studied \citep{arivazhagan-etal-2020-translation,sen-etal-2023-self}. 

 Sequence labelling is a staple of various incremental linguistic tasks possibly used in dialogue systems, like SRL \citep{konstas-etal-2014-incremental}, POS-tagging \citep{beuck-etal-2011-decision}, dialogue act segmentation \citep{manuvinakurike-etal-2016-toward}, disfluency detection \citep{Hough-2015} and dependency parsing \citep{honnibal-johnson-2014-joint}.

\paragraph{\textbf{Revision Categorisation and Prediction}} Approaches to categorise the properties of revisions or edits exist in various areas. \citet{faigley1981analyzing} examine the effects and causes of revisions in writing, providing a taxonomy on whether revisions change meaning and bring new information. \citet{afrin-litman-2018-annotation} classify revision quality by whether they improve student essays. \citet{anthonio-etal-2020-wikihowtoimprove} categorise revisions and edits in WikiHow in terms of what they cause to the text. Wikipedia's edits have also been classified according to factuality and fluency \citep{bronner-monz-2012-user} and intents \citep{rajagopal-etal-2022-one}. Other typologies and taxonomies have been proposed for translation revisions \citep{fujita-etal-2017-consistent} and multilingual NLG revision operations \citep{callaway-2003-multilingual}.

\citet{vaughan-mcdonald-1986-model} outline three phases of the revision process in NLG: Recognition, editing and re-generation. Revision rules have been applied for incremental summarisation by \citet{robin-1996-evaluating}. Non-incremental revision learning models also exist, relying on revision rules for dependency parsing \citep{attardi-ciaramita-2007-tree} or classification in POS-tagging \citep{nakagawa-etal-2002-revision}. Predicting stability and accuracy of hypotheses is a relevant task \citep{selfridge-etal-2011-stability}, which allows to distinguish hypotheses that will survive and are thus more reliable \citep{baumann-etal-2009-assessing}.

\paragraph{\textbf{Incremental Evaluation}} Table \ref{table:properties} presents an overview of relevant properties for incremental evaluation. In their seminal work, \citet{baumann-incremental} define three general categories of metrics for incremental processing: \textit{similarity}, \textit{timing} and \textit{diachronic}, which can be employed in incremental sequence labelling. They are suitable for capturing \textit{e.g.}~instability (edit overhead), quality of prefixes (correctness) and lag (correction time). \citet{kaushal-etal-2023-efficient} propose streaming exact match, comparing prefixes with the final gold standard. While these metrics capture instability and correctness of output prefixes, we lack a standard way to evaluate the quality of the performed revisions. We thus complement their work by proposing fine-grained metrics focusing on revisions and recomputations.

\section{Evaluation Methodology}
\label{sec:eval}
In this section, we present our evaluation methodology for incremental sequence labelling with a focus on revisions. After formalising the task, we characterise revisions and edits, define policies and revision-oriented metrics and discuss the ideal behaviour of incremental sequence labelling models.

\subsection{Formalisation}
\label{sec:formalisation}
We begin by formalising incremental sequence labelling tasks, extending the similar definition of streaming sequence tagging by \citet{kaushal-etal-2023-efficient} with \textit{edits} and \textit{revisions}. Like them, we assume an idealised format where incremental units are well-defined, fixed and complete input tokens, and a model that produces a label for every new input token, so that the output is necessarily extended at every time step. Note, however, that incremental processors may have to operate at sub-token level or with transitional input, which requires the capability of retracting decisions and adjusting to varying length in real-time. In some models, outputs may not have an immediate one-to-one correspondence to the input (\textit{e.g.}~due to a delay strategy \citep{baumann-incremental}, or to techniques like opportunistic decoding \citep{zheng-etal-2020-opportunistic}) and parallel hypotheses can be kept in memory. See \citet{iu-restart} for details.

Let $L=\{L_1, \ldots, L_M\}$ be a set of labels. In standard sequence labelling, the task is to map an input sequence of $n$ tokens $(w_i)_{i=1}^n$ to an output sequence of $n$ labels $(l_i)_{i=1}^n$, $l_i \in L$. Each output label $l_i$ classifies its corresponding token $w_i$. The task is more complex than plain token-level classification because the sequential nature of the input and the output need to be taken into account when predicting labels. If available, a gold-standard sequence $(g_i)_{i=1}^n$, with $g_i \in L$, is used to evaluate the correctness of the predicted output sequence.

\begin{table*}[ht!]
\centering
\small
\resizebox{\textwidth}{!}{%
\begin{tabular}{@{}rl|lc|lc@{}}
     \toprule        
                                & \textbf{Quality}  & \textbf{Edits} (labels)                            &  \textbf{Example}     & \textbf{Revisions} (prefixes)           & \textbf{Example} \\ \toprule
    \textbf{Convenience}        & convenient        & change incorrect label                    &  (5,1)                & change incorrect prefix               & 5   \\
                                & inconvenient      & change correct label                      &  (4,2)                & change correct prefix                 & 4   \\ \midrule
    \textbf{Effectiveness}      & effective         & incorrect label $\rightarrow$ correct     &  (5,4)                & improve prefix correctness            & 6   \\
                                & ineffective       & incorrect label $\rightarrow$ incorrect   &  (9,3)                & do not change prefix correctness      & 9   \\
                                & defective         & correct label $\rightarrow$ incorrect     &  (4,3)                & worsen prefix correctness             & 4   \\ \midrule
    \textbf{Novelty}            & innovative        & label $\rightarrow$ new state             &  (9,6)                & n/a                                   & n/a \\
                                & repetitive        & label $\rightarrow$ previous state        &  (6,1)                & n/a                                   & n/a \\ \midrule
\textbf{(Local) Recurrence}     & recurrent         & subsequence with $>1$ edit                &  (9,3)                & subsequence with $>1$ revision        & 8   \\
                                & steady            & subsequence with $1$ edit                 &  (4,2)                & subsequence with $1$ revision         & 2   \\ \midrule
    \textbf{Oscillation}        & oscillating       & label with $>1$ edit                      &  (6,1)                & $>1$ revision                         & all    \\
                                & stable            & label with $1$ edit                       &  (4,2)                & single revision                       & -    \\ \midrule
    \textbf{Company}            & accompanied       & prefix with $>1$ edit                     &  (9,6)                & prefix with $>1$ edit                 & 5   \\
                                & isolated          & prefix with $1$ edit                      &  (6,1)                & prefix with $1$ edit                  & 6   \\ \midrule
    \textbf{Connectedness}      & connected         & other neighbouring edit                   &  (9,4)                & only connected edits                  & 9   \\
                                & disconnected      & no neighbouring edits                     &  (5,1)                & only disconnected edits               & 2   \\
                                & both              & n/a                                       &  n/a                  & both types of edits                   & 5   \\ \midrule
    \textbf{Distance}           & short range       & near current time step                    &  (5,4)                & only short range edits                & 2   \\
                                & long range        & far from current time step                &  (9,3)                & only long range edits                 & 6   \\
                                & both              & n/a                                       &  n/a                  & both types of edits                   & 5   \\ \midrule
    \textbf{Definiteness}       & definite          & label $\rightarrow$ final state           &  (4,2)                & prefix $\rightarrow$ final state      & 10  \\
                                & temporary         & label $\rightarrow$ temporary state       &  (5,3)                & prefix $\rightarrow$ temporary state  & 8   \\ \midrule
    \textbf{Time}               & intermediate      & input still partial                       &  (5,4)                & input is still partial                & 4   \\
                                & final             & at final time step                        &  (10,3)               & at the final time step                & 10  \\ \bottomrule
\end{tabular}%
}
\caption{Characterisation of edits and revisions. The examples refer to Figure \ref{fig:contructed-example}, pointing to the (time step, label index) positions for edits and time steps for revisions. Here the gold standard is used to judge prefix correctness.}
\label{table:charac}
\end{table*}

In an incremental setting, the input is provided in a piecemeal fashion, one token at a time. At each time step $t=1, 2, \ldots, n$, an increasing input prefix $(w_i)_{i=1}^t$ is available to the model and an output prefix $(l_i)_{i=1}^t$ is predicted. Therefore, an input sequence with $n$ tokens will result in $n$ output prefixes $p_1, p_2, \ldots, p_n$, which we consider to be partial hypotheses for the final output. Each $p_i$ is a sequence of $i$ labels, containing one additional label at the right in relation to $p_{i-1}$. The last hypothesis $p_n$ is the final decision of the model, having observed the full input. The complete sequence of prefixes can be represented as a lower triangular matrix, whose cells $c_i^j$ contain the label assigned to $w_i$ at time $j$ and each row $i$ contains $p_i$. We can represent the incremental input and output in an \textit{incremental chart} (IC) as follows:

\vspace{0.2cm}
\begin{center} 
    $\begin{NiceArray}{r|r|ccccc} 

        \toprule    
        w_1                         & p_1=              & l_1^1     &           &           &           &         \\ 
        w_1, w_2                    & p_2=              & l_1^2     &   l_2^2   &           &           &         \\ 
        w_1, w_2, w_3               & p_3=              & l_1^3     &   l_2^3   &  l_3^3    &           &         \\ 
        \vdots                      & \vdots            & \vdots    &   \vdots  &  \vdots   &  \ddots   &         \\
        w_1, w_2, \ldots, w_n       & p_n=              & l_1^n     & l_2^n     &  l_3^n    & \cdots    & l_n^{n} \\ \cmidrule{2-7}
                                    & \text{gold} =     & g_1       & g_2       &  g_3      & \cdots    & g_n     \\  \bottomrule
    \end{NiceArray}$
\end{center}
\vspace{0.2cm}

At each time step $t$, the observation of the new input token $w_t$ causes the model to i) extend the output sequence with one label for $w_t$ (an addition) and ii) optionally also change its current hypotheses $l_1, \ldots, l_{t-1}$ for previous tokens (substitutions). 

An \textit{edit} occurs at time $t$ for label $i$ if $l_{i}^{t} \neq l_{i}^{t-1}$, meaning that the model's prediction for $w_i$'s label changed. A \textit{revision} occurs when, apart from the compulsory addition, a prefix changes at time $t$ in relation to the previous prefix, \textit{i.e.}~when at least one label is edited.\footnote{The addition is not taken into account here, as it has no precedent label to be compared to at this point. The first time step is by definition not a revision, since there is no prefix yet.}  In Figure \ref{fig:contructed-example}, revisions occur at time steps $2, 4, 5, 6, 8, 9$ and $10$. Highlighted labels in the prefixes are edits. 

\begin{figure}[t]
    \centering
    \includegraphics[trim={0cm 14cm 4cm 1cm},clip,width=0.8\columnwidth,page=1]{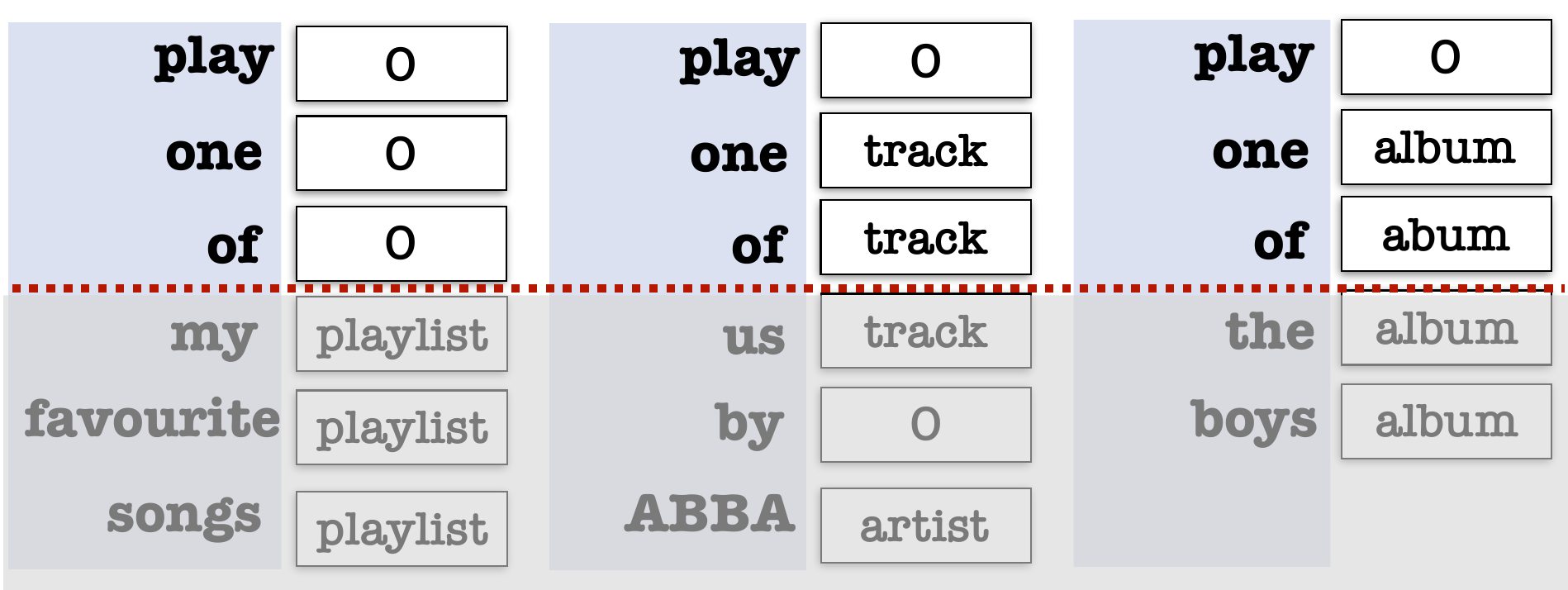}
    \caption{Illustrative example of multiple locally valid hypotheses for the prefix \textit{play one of}. Only after more input is processed definite labels can be assigned.}
    \label{fig:locally-valid}
\end{figure}

\paragraph{\textbf{Gold Standard}} Evaluation can be done with respect to incremental or non-incremental gold standards \citep{baumann-incremental}. Often, only the non-incremental version is available, \textit{i.e.}~the labels on the complete sequence, assigned having all left and right context taken into account. A genuinely incremental gold standard contains step-by-step gold prefixes encoding interpretations that are \textit{locally valid} until right context renders it invalid, as illustrated in Figure \ref{fig:locally-valid}.\footnote{For existing examples, see \citet{hrycyk-etal-2021-fast}, \citet{rawat2022real} and \citet{beuck-inc-parsing}.} Since it is usually not available, we can instead ``incrementalise'' the final gold standard by deriving all its prefixes as hard labels. But this approach somewhat unfairly expects that, even at steps with multiple locally valid interpretations, the model commits to the final decision without observing the input that actually induces that interpretation as correct and the others as wrong. Moreover, using an independent gold standard conflates the external overall performance of the model with the quality of its internal incrementality; an alternative is to consider the final output of the model as a silver standard \citep{baumann-incremental}. The correctness of labels and prefixes is then measured with a metric $M$ with respect to the defined target.

\subsection{Characterisation of Revisions and Edits}
\label{sec:characterisation}
In this section, we propose a detailed characterisation for the types of edits and revisions based on ten dimensions, summarised in Table \ref{table:charac}, as a means to evaluate revision policies. In the next paragraphs, we assume that either a genuine or a constructed incremental sequence of target prefixes has been selected according to the current needs. We will use Figure \ref{fig:contructed-example} and its gold standard as examples.\footnote{More examples are available in the code repository.}

To characterise edits, we consider the state of an output label in the current prefix in relation to its state in the previous prefix, which are different by definition. They relate to a label's development in time (vertically in their IC column) or to the prefix they belong to (horizontally in their IC row). The dimensions to characterise edits serve the purpose of defining the qualities of the revisions, which operate on prefixes.

\subsubsection{Edits}

The main aspect to account for is whether labels need to be edited in the first place and, if yes, whether they are edited into the desired state. Edits on correct labels are \textit{inconvenient}, and also \textit{defective}, since the label will fatally change into a wrong label. This happens, for instance, at $l_2$ in step $4$, as the correct label $D$ is edited into a wrong $B$. Edits on incorrect labels are \textit{convenient} and can be \textit{effective} (if it enters into a correct state, like $l_4$ at $t=5$, which changes from an incorrect $D$ to a correct $C$) or \textit{ineffective} (if it enters into another incorrect state, \textit{e.g.}~$l_3$ at $t=9$, which changed from an incorrect $A$ to a still incorrect $D$). 

Other dimensions can be used to analyse the behaviour of the processor. \textit{Innovative} edits cause the label to change into a new state. For instance, $l_6$ becomes a $C$ for the first time at $t=9$. In the next step, it is edited back into its previous state $A$, and we consider it to be a \textit{repetitive} edit.

Local \textit{recurrence} refers to whether the edit occurs in isolation in neighbouring time steps (edit subsequences in an IC's column). \textit{Oscillation} refers to how many edits occur in its complete column, just one (\textit{stable}) or more (\textit{oscillating}). For instance, $l_3$ has two groups of recurrent edits along the time axis, whereas $l_2$ has one \textit{steady} and stable edit. 

\textit{Company} characterises whether the edit occurs with other edits in a prefix (same IC's row). In Figure \ref{fig:contructed-example}, $l_6$ is edited together with other labels at $t=9$, whereas $l_1$ is edited in isolation at $t=6$. \textit{Accompanied} edits can be either \textit{connected} (\textit{i.e.}~with directly neighbouring edited labels, as in $t=4$) or \textit{disconnected} to the other edits in its prefix.

\textit{Short} or \textit{long range} refers to how far the edited label is from the current time step, defined by a distance parameter $d$. If we set $d=2$, the edit that changes $l_4$ into a $C$ at $t=5$ is short range because it is less than 2 time steps away from the current token being processed. On the other hand, $l_3$ is edited at $t=9$, very distant from the right frontier.

Edits can also be \textit{definite} or \textit{temporary}. Definite edits make the label enter into its final state, like $l_2$ at $t=4$. Temporary edits are those like the $B$ for $l_3$ at $t=5$: It still gets edited further before a final decision is reached (here, also a $B$). Besides, edits can occur in \textit{intermediate} steps during processing, when the input sequence is incomplete, or at the \textit{final} time step, when the full sequence is available.

\subsubsection{Revisions}

 Similar to edits, revisions are \textit{inconvenient} if they occur on correct prefixes (that should not change), and thus also \textit{defective}, because correctness necessarily decreases. The prefix at $t=3$ is correct, so the revision at $t=4$ causes the labels to become wrong. \textit{Convenient} revisions are \textit{effective} if they improve correctness, like at $t=6$ where the number of correct labels in the prefix increases from $3$ to $4$, otherwise they can be \textit{ineffective} (edits occur but correctness remains the same, like at $t=9$) or again \textit{defective}.

Revisions are \textit{locally recurrent} when other revisions occur in neighbouring time steps. We see that from $t=4$ to $t=6$. The revision at $t=2$ is \textit{steady}, as no other revisions occur immediately before or after it. If only one revision occur while a sequence is processed, it is \textit{stable}, otherwise it is \textit{oscillating.} In our example, all revisions are therefore oscillating.

\textit{Company}, \textit{connectedness} and \textit{distance} refer to what types of edits the revision causes. At the second time step, the prefix contains only a \textit{disconnected} and \textit{short range} edit, whereas at the fifth time step we observe \textit{accompanied} edits, one \textit{connected} and one \textit{disconnected} group and one short and two long range edits.

\textit{Definite} revisions create prefixes that will not be further edited. In our example, this only happens in the last time step; all others are \textit{temporary}. \textit{Intermediate} revisions happen when the input is not yet completed, otherwise they are \textit{final}.

\subsubsection{Recomputations}

In models that detach recomputations from revisions, the recomputations should also be evaluated. Recomputations are \textit{active} if they actually result in a revision, otherwise they are \textit{inactive}. The quality of the resulting revisions can then be evaluated with the characteristics above.

\subsection{Policies}
\label{sec:policies}
To perform good revisions, a model must decide \textit{when} to recompute or revise. For that decision, both a \textit{revision policy} and a \textit{recomputation policy} can be generally defined as:

\begin{equation}
    \pi: \text{IC} \rightarrow [0,1] \hspace{0.8cm} \pi(\text{IC}_t) = \Pr(r | \text{IC}_t )
\end{equation}

It gives the probability of performing a revision or recomputation $r$, respectively, given the state of the incremental chart at time $t$.\footnote{It is also possible to make the policy dependent only in a portion of the $IC$, as done \textit{e.g.}~by \citet{tapir}.} When $\Pr(r | \text{IC}_t ) > \tau$, where $\tau$ is a threshold hyperparameter, a revision/recomputation is performed. If the revisions are not a mere consequence of full recomputations, the model must then also decide \textit{what} and \textit{how} to edit.

\subsection{Metrics}
\label{sec:metrics}
\begin{table*}[t!]
    \centering
    \small
    \begin{tabular}{rcl}
    \toprule
        & & The fraction of... \\
        \textbf{Rate of Revision} & $R/N$ & \hspace{1cm} time steps in which the model revises \\
        \textbf{Rate of Recomputation} & $R'/N$ & \hspace{1cm}  time steps in which the model recomputes \\
        \textbf{Rate of Active Recomputation} & $(R'\cap R)/R'$ & \hspace{1cm} recomputations that actually causes a revision \\
        \midrule
        \textbf{R-Pertinence} & $(R \cap I) / R$ & \hspace{1cm}  revisions that edit incorrect prefixes (adapted precision)  \\
        \textbf{R-Appropriateness} & $(R \cap I) / I$ & \hspace{1cm} incorrect prefixes that are revised (adapted recall) \\
        \textbf{A-Pertinence} & $(A \cap C) / A$ & \hspace{1cm}  additions upon correct prefixes (adapted precision) \\
        \textbf{A-Appropriateness} & $(A \cap C) / C$ & \hspace{1cm}  correct prefixes that are not revised (adapted recall) \\
        \midrule
        \textbf{R$_e$-Pertinence} & $(R_e \cap I) / R$ & \hspace{1cm}  revisions that effectively edit incorrect prefixes   \\
        \textbf{R$_e$-Appropriateness} & $(R_e \cap I) / I$ & \hspace{1cm} incorrect prefixes that are revised effectively \\
    \bottomrule
    \end{tabular}

    \caption{Proposed metrics for evaluating recomputation and revision policies. $N$ is the total number of time steps.}
    \label{table:metrics}
\end{table*}

Traditional sequence labelling evaluation metrics like accuracy or F1 can be computed on label, sequence or dataset level. The incremental dimension requires its own metrics, some of which we discussed in $\S$\ref{sec:litreview}. Here, we propose specific metrics to evaluate revision and/or recomputation policies. For each time step $t$ in a sequence, either a revision ($R$) occurred, which is sometimes effective ($R_{e}$), or only an addition ($A$). Assuming we have established a metric for prefix correctness,\footnote{A binary variable or a continuous variable, like accuracy, with a defined threshold for tolerated incorrectness.} we know whether the prefix at $t-1$ was correct ($C$) or incorrect ($I$). That results in a distribution of $N$ actions in $\{R, A\} \times \{C, I\}$. From these counts, we derive the metrics in Table \ref{table:metrics}, computed either per sequence or over the whole dataset. Models that have the option to \textit{recompute} ($R'$) can also be evaluated in $\{R', \neg R'\} \times \{C, I\}$ with two additional metrics. 

Since only \textit{effective} revisions are actually desired, the $R$ in the numerators can be replaced by $R_{e}$ for a more focused evaluation. Revisions can be further weighted by how often and how far in the sentence processing they happen. Similarly, edits can be assessed by their correction time and survival time \citep{baumann2013:phd}.

\begin{table*}[ht!]
    \centering
    \small

    \begin{tabular}{rrrrrrrrrr}
    \toprule
         & \multicolumn{3}{c}{\textbf{\% recomputation}} & \multicolumn{3}{c}{\textbf{\% active recomputation}} & \multicolumn{3}{c}{\textbf{\% revision}} \\
    \cmidrule(lr){2-4} \cmidrule(lr){5-7} \cmidrule(lr){8-10}
            & NER & POS & Slot & NER & POS & Slot & NER & POS & Slot \\
    \cmidrule(lr){2-4} \cmidrule(lr){5-7} \cmidrule(lr){8-10}
        Rest.Incremental-Transformer & 100.00 & 100.00 & 100.00 & 7.77 & 19.29 & 21.23 & 7.77 & 19.29 & 21.23 \\
        \textsc{Tapir}-LTReviser & 13.77 & 24.52 & 20.34 & 20.23 & 39.55 & 39.44 & 2.78 & 9.69 & 8.02 \\
        \textsc{Tapir}-TrfReviser & 10.36 & 20.23 & 21.41 & 25.36 & 34.09 & 33.65 & 2.62 & 6.89 & 7.20 \\
    \bottomrule
    \end{tabular}

    \caption{Rate of (active) recomputations and of revisions for each model and task.}
    \label{table:rates}
\end{table*}

\subsection{Ideal Processor}
\label{sec:ideal}
Let us now delineate the ideal behaviour of a revision policy for an incremental sequence labelling model. A utopian model would always output the correct label and thus never need to produce edits or revisions \citep{tapir}.\footnote{That is indeed the case for strictly monotonic models if we use their final output as gold standard.} But due to the incremental nature of language processing, models should not be penalised for building hypotheses that are \textit{locally valid}, as long as a revision is timely triggered. That is, however, complex to know in raw textual input where local ambiguities are not identified. Instead, we can characterise an outlook according to desirable principles and available resources. In scenarios with an infinite time budget, we can simply wait for the input to be complete. If computation budget can be afforded, restart-incrementality is a good fit. But the constraints are not always so loose.

An ideal revision policy should thus revise as rarely as possible for stability. If a prefix/label is correct, the policy should avoid revising it, whereas an incorrect prefix/label should be revised (maybe not immediately, but eventually).
It should always trigger effective, convenient, and definite revisions, preferably in earlier time steps.\footnote{In the beginning, the absence of both right and left context makes prediction harder. Towards the end, the availability of more left context should lead to less, and better, revisions.} Recurrent or oscillating revisions cause more instability and should be avoided. Innovative edits are preferable (as long as they are effective), and short range is better to be combined with delay strategies. Connectedness is a relevant dimension for BIO labelling schemes: If, for instance, the beginning label is edited, ideally the middle labels should change simultaneously. Finally, accompanied edits can be further evaluated in their relation to each other and the linguistic input. A good recomputation policy should, additionally, always result in active revisions.  

In terms of metrics, R-Pertinence and A-Appropriateness should be exactly $1$, \textit{i.e.}~all revisions should occur upon incorrect prefixes and all correct prefixes should not be revised. A-Pertinence and R-Appropriateness should be as high as possible, but cannot be expected to be exactly $1$ because it may take some time steps until the input that actually resolves the ambiguity or mistake is observed.

\begin{figure*}[th!]
    \centering
    \includegraphics[trim={0cm 0cm 2cm 0cm},clip,width=\textwidth,page=1]{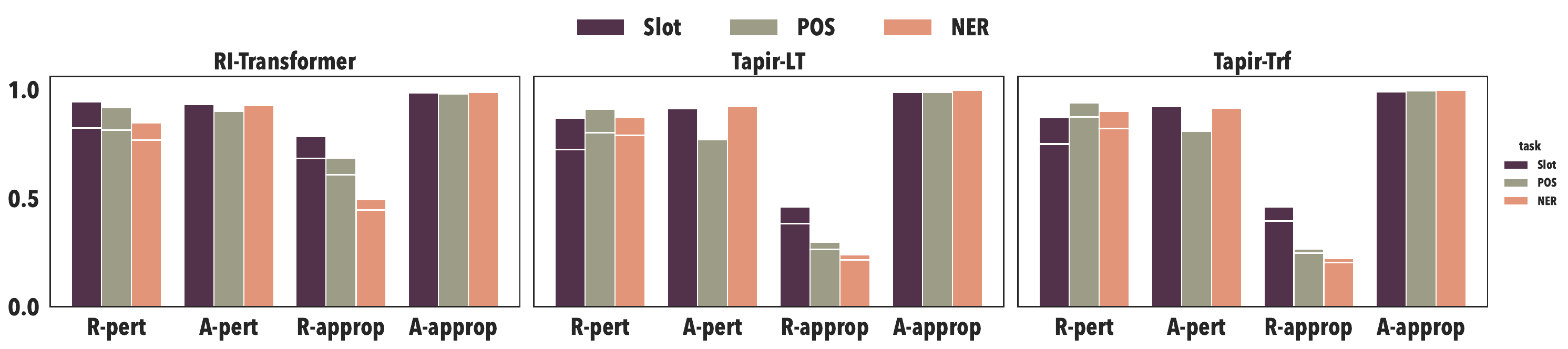}
    \caption{Revision metrics for all models and tasks. The white lines represent only the effective revisions.}
    \label{fig:analysis-metrics}
\end{figure*}

\section{Architecture Profiling}
\label{sec:analysis}
We now apply our methodology to profile the revision policy behaviour of three models: The reference restart-incremental Transformer and the two \textsc{Tapir} variations, which have a recomputation policy, proposed by \citet{tapir}. We evaluate them on three sequence labelling tasks: Slot filling \citep{coucke2018snips}, POS tagging \citep{silveira-etal-2014-gold} and NER \citep{tjong-kim-sang-de-meulder-2003-introduction}, using the final output as gold standard.\footnote{Here we use only the buffer outputs to evaluate the resulting revisions on prefixes that would have been passed on to downstream processors. We do not consider the temporary outputs of the LSTM that the original model had access to when deciding to perform a recomputation. Please refer to the original paper for the details on non-incremental and incremental performance on these tasks.} Note that the same profiling can be applied to any model with the ability of performing revisions on any sequence labelling task.

\paragraph{Quantitative Assessment} Table \ref{table:rates} shows that the recomputation policy implemented in \textsc{Tapir} reduces the number of restarts to between $10\%$ and $25\%$ in comparison to the restart incremental approach, considerably alleviating the computation load; the number of revisions is also 2 to 3 times lower. Still, only up to 40\% of the remaining recomputations are active, which means that the use of computational budget is still suboptimal. Furthermore, in Figure \ref{fig:analysis-metrics} we see that A-Appropriateness is very close to 1, as it should be. R-Pertinence is slightly below the ideal 1, but still greater than 0.8 in all cases, although it is around 0.1 lower when only effective revisions are considered. A-Pertinence is at similar values, with a lower result for POS-tagging. R-Appropriateness and R$_e$-appropriateness, however, are low in the restart-incremental Transformer and becomes even lower in the \textsc{Tapir} models. 

\begin{figure}[h!]
    \centering
        \includegraphics[trim={0cm 0.65cm 0cm 0cm},clip,width=\columnwidth,page=1]{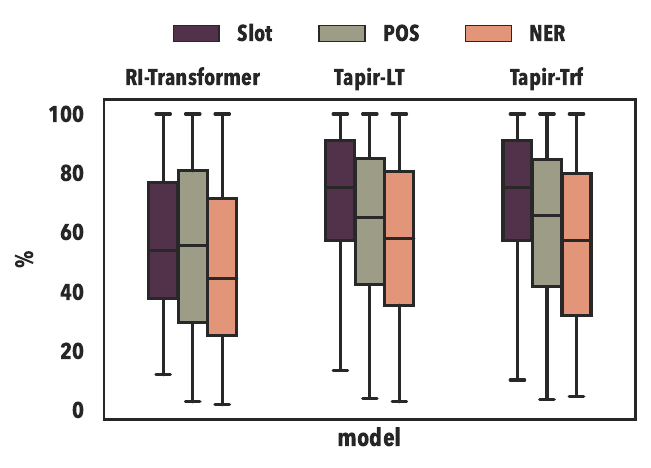}
        \caption{How far in the sentence processing (\% of time steps or tokens) revisions occur.}
        \label{fig:revision-when}
\end{figure}

This may be evidence that the \textsc{Tapir} models are waiting for more input before deciding to recompute an incorrect prefix, which is in line with the shifts in the distributions we observe in Figure \ref{fig:revision-when}. \textsc{Tapir} tends to have more revisions towards the end of the sentence than the restart-incremental Transformer. This strategy can indeed help revisions be more effective, given that more left context is available, but it also results in having to wait longer for final decisions, which is not ideal. 

\begin{figure}
        \includegraphics[trim={0cm 0.7cm 2.09cm 0cm},clip,width=\columnwidth,page=1]{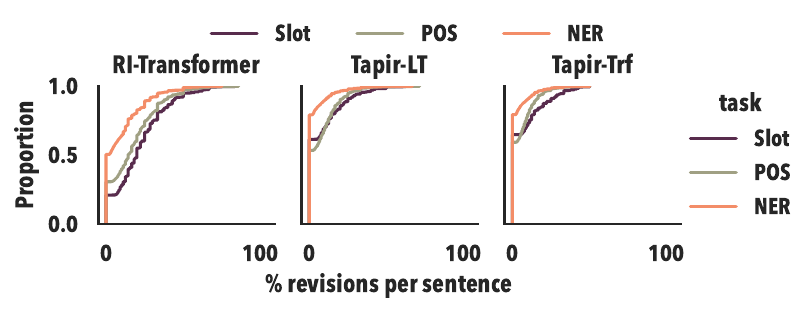}
        \caption{Proportion of time steps with revisions per sentence (cumulative).}
    \label{fig:revision-freq}   
\end{figure}

The cumulative distributions of the fraction of time steps with revisions per sentence, shown in Figure \ref{fig:revision-freq}, illustrate that the policy reduces the number of revisions per sentence: 50\% or less of the sentences have no revisions in the naive policy, which makes all recomputation effort be used to perform only an addition, while \textsc{Tapir}'s policy caused more sentences to not trigger revisions.

\begin{figure}[h!]
    \centering
    \includegraphics[trim={0.2cm 0.7cm 0cm 0cm},clip,width=\columnwidth,page=1]{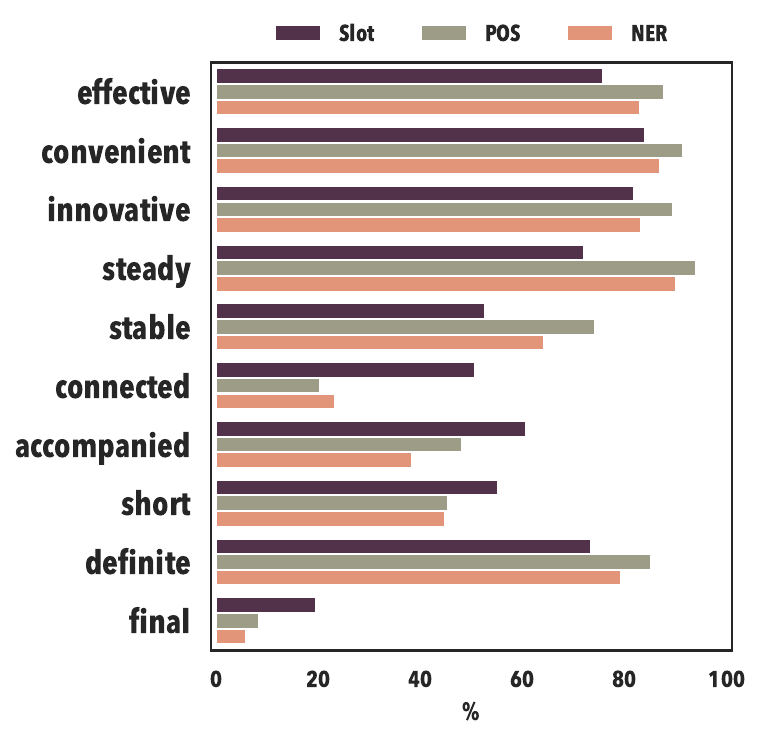}
    \caption{Edits by \textsc{Tapir}-TrfReviser's policy.}
    \label{fig:edit-types}
\end{figure}

\begin{figure}[h!]
    \centering
    \includegraphics[trim={0cm 0.7cm 0cm 0cm},clip,width=\columnwidth,page=1]{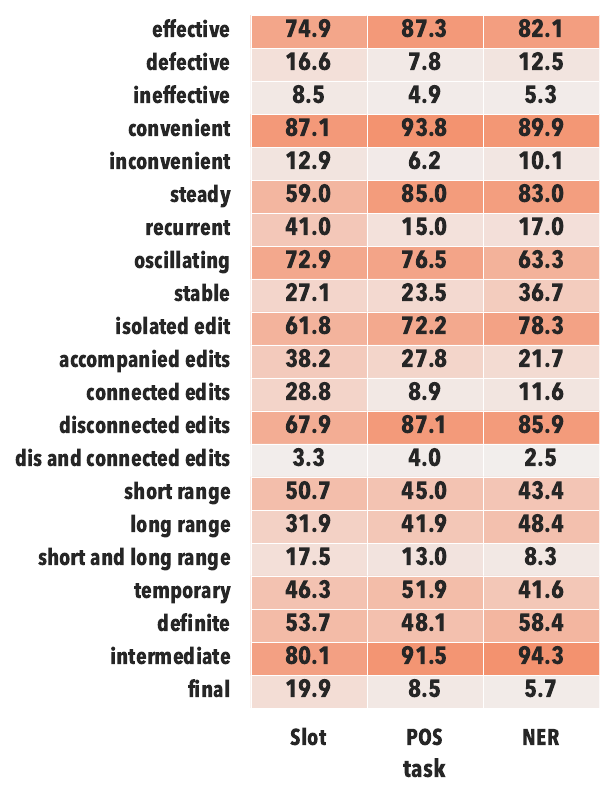}
    \caption{Revisions by \textsc{Tapir}-TrfReviser's policy.}
    \label{fig:revision-types}
\end{figure}

\paragraph{Qualitative Assessment} Figures \ref{fig:edit-types} and \ref{fig:revision-types} show the percentages of edits and revisions types to characterise \textsc{Tapir}-TrfReviser's policy. In terms of edits, most are effective, convenient, innovative and steady. Only around 50\% are short range, which means that delay strategies would have limited improvements in reducing edit overhead. For slot filling, around 20\% of the edits occur in the last time step, which is undesired, because it means that the intermediate predictions for these labels are wrong until the model processes the full sentence.

Regarding revisions, \textsc{Tapir}'s policy works best for POS-tagging in terms of effectiveness, convenience, oscillation and recurrence, and worse for slot filling. Most of the edits are isolated, which means that recomputations have been performed for the full partial input to only result in one edit. The proportion of short vs. long range and temporary vs. definite revisions was, in general, balanced. We also see that proportionally fewer revisions occurred in the final step. Although the high percentage of intermediate revisions is high, Figure \ref{fig:revision-when} shows that they are happening towards the end, which prevents incremental subprocessors to reliably count on the intermediate outputs. Slot filling is, here, an example of the occurrence of final revisions being less than ideal.

Based on these results, we conclude that \textsc{Tapir}'s policy is very successful in reducing the number of recomputations and also in revising less, but there is room for improving the quality of the resulting revisions, both in terms of metrics and of characteristics. This speaks for a more dedicated revision policy that could avoid full recomputations and use the state of the incremental chart and internal representations of the model for a more fine-grained prediction of which labels should change.

\section{Conclusion}
\label{sec:conclusion}
In this work, we have argued that the importance of a solid evaluation framework for revision policies in incremental sequence labelling cannot be overstated. Despite being very useful to capture some incremental aspects like instability or timeliness, existing evaluation metrics set aside other major strands of revisions. To fill that void, we have introduced metrics, characteristics and rationale to support the analysis of revision policies. This methodology serves as a tool to ascertain their quality, to determine their appropriateness in different contexts and to compare different policies. 

We identify a few more roads to quality: The creation of incremental gold standards containing locally valid hypothesis, the development of fine-grained revision policies predicting what to revise and a more systematic integration of linguistic aspects of the input into the evaluation procedure. For those willing to drive those routes, we hope our methodology has paved the road well. 

\section*{Acknowledgements}
We thank the anonymous reviewers for their valuable comments and suggestions. We also thank \citet{kaushal-etal-2023-efficient} for a conversation on this topic at EACL, in particular about the locally valid hypotheses.

\bibliography{anthology,custom}

\begin{thebibliography}{51}
\expandafter\ifx\csname natexlab\endcsname\relax\def\natexlab#1{#1}\fi

\bibitem[{Addlesee et~al.(2020)Addlesee, Yu, and
  Eshghi}]{addlesee-etal-2020-comprehensive}
Angus Addlesee, Yanchao Yu, and Arash Eshghi. 2020.
\newblock \href {https://doi.org/10.18653/v1/2020.coling-main.312} {A
  comprehensive evaluation of incremental speech recognition and diarization
  for conversational {AI}}.
\newblock In \emph{Proceedings of the 28th International Conference on
  Computational Linguistics}, pages 3492--3503, Barcelona, Spain (Online).
  International Committee on Computational Linguistics.

\bibitem[{Afrin and Litman(2018)}]{afrin-litman-2018-annotation}
Tazin Afrin and Diane Litman. 2018.
\newblock \href {https://doi.org/10.18653/v1/W18-0528} {Annotation and
  classification of sentence-level revision improvement}.
\newblock In \emph{Proceedings of the Thirteenth Workshop on Innovative Use of
  {NLP} for Building Educational Applications}, pages 240--246, New Orleans,
  Louisiana. Association for Computational Linguistics.

\bibitem[{Anthonio et~al.(2020)Anthonio, Bhat, and
  Roth}]{anthonio-etal-2020-wikihowtoimprove}
Talita Anthonio, Irshad Bhat, and Michael Roth. 2020.
\newblock \href {https://aclanthology.org/2020.lrec-1.702}
  {wiki{H}ow{T}o{I}mprove: A resource and analyses on edits in instructional
  texts}.
\newblock In \emph{Proceedings of the Twelfth Language Resources and Evaluation
  Conference}, pages 5721--5729, Marseille, France. European Language Resources
  Association.

\bibitem[{Arivazhagan et~al.(2020)Arivazhagan, Cherry, Macherey, and
  Foster}]{arivazhagan-etal-2020-translation}
Naveen Arivazhagan, Colin Cherry, Wolfgang Macherey, and George Foster. 2020.
\newblock \href {https://doi.org/10.18653/v1/2020.iwslt-1.27} {Re-translation
  versus streaming for simultaneous translation}.
\newblock In \emph{Proceedings of the 17th International Conference on Spoken
  Language Translation}, pages 220--227, Online. Association for Computational
  Linguistics.

\bibitem[{Attardi and Ciaramita(2007)}]{attardi-ciaramita-2007-tree}
Giuseppe Attardi and Massimiliano Ciaramita. 2007.
\newblock \href {https://aclanthology.org/N07-1049} {Tree revision learning for
  dependency parsing}.
\newblock In \emph{Human Language Technologies 2007: The Conference of the
  North {A}merican Chapter of the Association for Computational Linguistics;
  Proceedings of the Main Conference}, pages 388--395, Rochester, New York.
  Association for Computational Linguistics.

\bibitem[{Baumann(2013)}]{baumann2013:phd}
Timo Baumann. 2013.
\newblock \href {http://arxiv.org/abs/urn:nbn:de:hbz:361-25819101}
  {\emph{Incremental Spoken Dialogue Processing: Architecture and Lower-level
  Components}}.
\newblock Ph.D. thesis, Universität Bielefeld, Germany.

\bibitem[{Baumann et~al.(2009)Baumann, Atterer, and
  Schlangen}]{baumann-etal-2009-assessing}
Timo Baumann, Michaela Atterer, and David Schlangen. 2009.
\newblock \href {https://aclanthology.org/N09-1043} {Assessing and improving
  the performance of speech recognition for incremental systems}.
\newblock In \emph{Proceedings of Human Language Technologies: The 2009 Annual
  Conference of the North {A}merican Chapter of the Association for
  Computational Linguistics}, pages 380--388, Boulder, Colorado. Association
  for Computational Linguistics.

\bibitem[{Baumann et~al.(2014)Baumann, Bangalore, and
  Hirschberg}]{baumann-etal-2014-towards}
Timo Baumann, Srinivas Bangalore, and Julia Hirschberg. 2014.
\newblock \href {https://aclanthology.org/2014.iwslt-papers.2} {Towards
  simultaneous interpreting: the timing of incremental machine translation and
  speech synthesis}.
\newblock In \emph{Proceedings of the 11th International Workshop on Spoken
  Language Translation: Papers}, pages 163--168, Lake Tahoe, California.

\bibitem[{Baumann et~al.(2011)Baumann, Buß, and
  Schlangen}]{baumann-incremental}
Timo Baumann, Okko Buß, and David Schlangen. 2011.
\newblock \href {https://doi.org/10.5087/dad.2011.106} {{Evaluation and
  Optimisation of Incremental Processors}}.
\newblock \emph{Dialogue and Discourse}, 2(1):113--141.

\bibitem[{Beuck et~al.(2011{\natexlab{a}})Beuck, K{\"o}hn, and
  Menzel}]{beuck-etal-2011-decision}
Niels Beuck, Arne K{\"o}hn, and Wolfgang Menzel. 2011{\natexlab{a}}.
\newblock \href {https://aclanthology.org/W11-4605} {Decision strategies for
  incremental {POS} tagging}.
\newblock In \emph{Proceedings of the 18th Nordic Conference of Computational
  Linguistics ({NODALIDA} 2011)}, pages 26--33, Riga, Latvia. Northern European
  Association for Language Technology (NEALT).

\bibitem[{Beuck et~al.(2013)Beuck, K{\"o}hn, and Menzel}]{beuck2013predictive}
Niels Beuck, Arne K{\"o}hn, and Wolfgang Menzel. 2013.
\newblock Predictive incremental parsing and its evaluation.
\newblock In \emph{Computational Dependency Theory}, pages 186--206. IOS Press.

\bibitem[{Beuck et~al.(2011{\natexlab{b}})Beuck, Köhn, and
  Menzel}]{beuck-inc-parsing}
Niels Beuck, Arne Köhn, and Wolfgang Menzel. 2011{\natexlab{b}}.
\newblock \href
  {https://edoc.sub.uni-hamburg.de/informatik/volltexte/2014/194/} {Incremental
  parsing and the evaluation of partial dependency analyses}.
\newblock In \emph{Proceedings of the 1st International Conference on
  Dependency Linguistics}, pages 290--299.

\bibitem[{Bronner and Monz(2012)}]{bronner-monz-2012-user}
Amit Bronner and Christof Monz. 2012.
\newblock \href {https://aclanthology.org/E12-1036} {User edits classification
  using document revision histories}.
\newblock In \emph{Proceedings of the 13th Conference of the {E}uropean Chapter
  of the Association for Computational Linguistics}, pages 356--366, Avignon,
  France. Association for Computational Linguistics.

\bibitem[{Callaway(2003)}]{callaway-2003-multilingual}
Charles Callaway. 2003.
\newblock \href {https://aclanthology.org/W03-2303} {Multilingual revision}.
\newblock In \emph{Proceedings of the 9th {E}uropean Workshop on Natural
  Language Generation ({ENLG}-2003) at {EACL} 2003}, Budapest, Hungary.
  Association for Computational Linguistics.

\bibitem[{Chen et~al.(2022)Chen, Zayats, Walker, and
  Padfield}]{chen-etal-2022-teaching}
Angelica Chen, Vicky Zayats, Daniel Walker, and Dirk Padfield. 2022.
\newblock \href {https://doi.org/10.18653/v1/2022.naacl-main.60} {Teaching
  {BERT} to wait: Balancing accuracy and latency for streaming disfluency
  detection}.
\newblock In \emph{Proceedings of the 2022 Conference of the North American
  Chapter of the Association for Computational Linguistics: Human Language
  Technologies}, pages 827--838, Seattle, United States. Association for
  Computational Linguistics.

\bibitem[{Coucke et~al.(2018)Coucke, Saade, Ball, Bluche, Caulier, Leroy,
  Doumouro, Gisselbrecht, Caltagirone, Lavril, Primet, and
  Dureau}]{coucke2018snips}
Alice Coucke, Alaa Saade, Adrien Ball, Théodore Bluche, Alexandre Caulier,
  David Leroy, Clément Doumouro, Thibault Gisselbrecht, Francesco Caltagirone,
  Thibaut Lavril, Maël Primet, and Joseph Dureau. 2018.
\newblock \href {http://arxiv.org/abs/1805.10190} {Snips voice platform: an
  embedded spoken language understanding system for private-by-design voice
  interfaces}.
\newblock \emph{arXiv preprint}, arXiv:1805.10190.

\bibitem[{Faigley and Witte(1981)}]{faigley1981analyzing}
Lester Faigley and Stephen Witte. 1981.
\newblock Analyzing revision.
\newblock \emph{College composition and communication}, 32(4):400--414.

\bibitem[{Fujita et~al.(2017)Fujita, Tanabe, Toyoshima, Yamamoto, Kageura, and
  Hartley}]{fujita-etal-2017-consistent}
Atsushi Fujita, Kikuko Tanabe, Chiho Toyoshima, Mayuka Yamamoto, Kyo Kageura,
  and Anthony Hartley. 2017.
\newblock \href {https://doi.org/10.18653/v1/W17-0807} {Consistent
  classification of translation revisions: A case study of {E}nglish-{J}apanese
  student translations}.
\newblock In \emph{Proceedings of the 11th Linguistic Annotation Workshop},
  pages 57--66, Valencia, Spain. Association for Computational Linguistics.

\bibitem[{Grenander et~al.(2022)Grenander, Cohen, and
  Steedman}]{grenander-etal-2022-sentence}
Matt Grenander, Shay~B. Cohen, and Mark Steedman. 2022.
\newblock \href {https://aclanthology.org/2022.emnlp-main.28}
  {Sentence-incremental neural coreference resolution}.
\newblock In \emph{Proceedings of the 2022 Conference on Empirical Methods in
  Natural Language Processing}, pages 427--443, Abu Dhabi, United Arab
  Emirates. Association for Computational Linguistics.

\bibitem[{Hochreiter and Schmidhuber(1997)}]{lstm}
Sepp Hochreiter and J\"{u}rgen Schmidhuber. 1997.
\newblock \href {https://doi.org/10.1162/neco.1997.9.8.1735} {Long short-term
  memory}.
\newblock \emph{Neural Comput.}, 9(8):1735–1780.

\bibitem[{Honnibal and Johnson(2014)}]{honnibal-johnson-2014-joint}
Matthew Honnibal and Mark Johnson. 2014.
\newblock \href {https://doi.org/10.1162/tacl_a_00171} {Joint incremental
  disfluency detection and dependency parsing}.
\newblock \emph{Transactions of the Association for Computational Linguistics},
  2:131--142.

\bibitem[{Hough and Schlangen(2015)}]{Hough-2015}
Julian Hough and David Schlangen. 2015.
\newblock {Recurrent Neural Networks for Incremental Disfluency Detection}.
\newblock In \emph{Interspeech 2015}, pages 849--853.

\bibitem[{Hrycyk et~al.(2021)Hrycyk, Zarcone, and Hahn}]{hrycyk-etal-2021-fast}
Lianna Hrycyk, Alessandra Zarcone, and Luzian Hahn. 2021.
\newblock \href {https://doi.org/10.18653/v1/2021.nlp4convai-1.6} {Not so fast,
  classifier {--} accuracy and entropy reduction in incremental intent
  classification}.
\newblock In \emph{Proceedings of the 3rd Workshop on Natural Language
  Processing for Conversational AI}, pages 52--67, Online. Association for
  Computational Linguistics.

\bibitem[{Kahardipraja et~al.(2021)Kahardipraja, Madureira, and
  Schlangen}]{kahardipraja-etal-2021-towards}
Patrick Kahardipraja, Brielen Madureira, and David Schlangen. 2021.
\newblock \href {https://doi.org/10.18653/v1/2021.emnlp-main.90} {Towards
  incremental transformers: An empirical analysis of transformer models for
  incremental {NLU}}.
\newblock In \emph{Proceedings of the 2021 Conference on Empirical Methods in
  Natural Language Processing}, pages 1178--1189, Online and Punta Cana,
  Dominican Republic. Association for Computational Linguistics.

\bibitem[{Kahardipraja et~al.(2023)Kahardipraja, Madureira, and
  Schlangen}]{tapir}
Patrick Kahardipraja, Brielen Madureira, and David Schlangen. 2023.
\newblock \href {https://aclanthology.org/2023.findings-acl.257} {{TAPIR}:
  Learning adaptive revision for incremental natural language understanding
  with a two-pass model}.
\newblock In \emph{Findings of the Association for Computational Linguistics:
  ACL 2023}, pages 4173--4197, Toronto, Canada. Association for Computational
  Linguistics.

\bibitem[{Kaushal et~al.(2023)Kaushal, Gupta, Upadhyay, and
  Faruqui}]{kaushal-etal-2023-efficient}
Ayush Kaushal, Aditya Gupta, Shyam Upadhyay, and Manaal Faruqui. 2023.
\newblock \href {https://aclanthology.org/2023.eacl-main.31} {Efficient
  encoders for streaming sequence tagging}.
\newblock In \emph{Proceedings of the 17th Conference of the European Chapter
  of the Association for Computational Linguistics}, pages 418--429, Dubrovnik,
  Croatia. Association for Computational Linguistics.

\bibitem[{K{\"o}hn(2018)}]{kohn-2018-incremental}
Arne K{\"o}hn. 2018.
\newblock \href {https://aclanthology.org/C18-1253} {Incremental natural
  language processing: Challenges, strategies, and evaluation}.
\newblock In \emph{Proceedings of the 27th International Conference on
  Computational Linguistics}, pages 2990--3003, Santa Fe, New Mexico, USA.
  Association for Computational Linguistics.

\bibitem[{K{\"o}hn and Menzel(2014)}]{kohn-menzel-2014-incremental}
Arne K{\"o}hn and Wolfgang Menzel. 2014.
\newblock \href {https://doi.org/10.3115/v1/P14-2130} {Incremental predictive
  parsing with {T}urbo{P}arser}.
\newblock In \emph{Proceedings of the 52nd Annual Meeting of the Association
  for Computational Linguistics (Volume 2: Short Papers)}, pages 803--808,
  Baltimore, Maryland. Association for Computational Linguistics.

\bibitem[{Konstas et~al.(2014)Konstas, Keller, Demberg, and
  Lapata}]{konstas-etal-2014-incremental}
Ioannis Konstas, Frank Keller, Vera Demberg, and Mirella Lapata. 2014.
\newblock \href {https://doi.org/10.3115/v1/D14-1036} {Incremental semantic
  role labeling with {T}ree {A}djoining {G}rammar}.
\newblock In \emph{Proceedings of the 2014 Conference on Empirical Methods in
  Natural Language Processing ({EMNLP})}, pages 301--312, Doha, Qatar.
  Association for Computational Linguistics.

\bibitem[{Ma et~al.(2020)Ma, Dousti, Wang, Gu, and
  Pino}]{ma-etal-2020-simuleval}
Xutai Ma, Mohammad~Javad Dousti, Changhan Wang, Jiatao Gu, and Juan Pino. 2020.
\newblock \href {https://doi.org/10.18653/v1/2020.emnlp-demos.19} {{SIMULEVAL}:
  An evaluation toolkit for simultaneous translation}.
\newblock In \emph{Proceedings of the 2020 Conference on Empirical Methods in
  Natural Language Processing: System Demonstrations}, pages 144--150, Online.
  Association for Computational Linguistics.

\bibitem[{Madureira and Schlangen(2020)}]{madureira-schlangen-2020-incremental}
Brielen Madureira and David Schlangen. 2020.
\newblock \href {https://doi.org/10.18653/v1/2020.emnlp-main.26} {Incremental
  processing in the age of non-incremental encoders: An empirical assessment of
  bidirectional models for incremental {NLU}}.
\newblock In \emph{Proceedings of the 2020 Conference on Empirical Methods in
  Natural Language Processing (EMNLP)}, pages 357--374, Online. Association for
  Computational Linguistics.

\bibitem[{Manuvinakurike et~al.(2016)Manuvinakurike, Paetzel, Qu, Schlangen,
  and DeVault}]{manuvinakurike-etal-2016-toward}
Ramesh Manuvinakurike, Maike Paetzel, Cheng Qu, David Schlangen, and David
  DeVault. 2016.
\newblock \href {https://doi.org/10.18653/v1/W16-3632} {Toward incremental
  dialogue act segmentation in fast-paced interactive dialogue systems}.
\newblock In \emph{Proceedings of the 17th Annual Meeting of the Special
  Interest Group on Discourse and Dialogue}, pages 252--262, Los Angeles.
  Association for Computational Linguistics.

\bibitem[{Nakagawa et~al.(2002)Nakagawa, Kudo, and
  Matsumoto}]{nakagawa-etal-2002-revision}
Tetsuji Nakagawa, Taku Kudo, and Yuji Matsumoto. 2002.
\newblock \href {https://doi.org/10.3115/1073083.1073167} {Revision learning
  and its application to part-of-speech tagging}.
\newblock In \emph{Proceedings of the 40th Annual Meeting of the Association
  for Computational Linguistics}, pages 497--504, Philadelphia, Pennsylvania,
  USA. Association for Computational Linguistics.

\bibitem[{Pozzan and Trueswell(2015)}]{pozzan2015revise}
Lucia Pozzan and John~C Trueswell. 2015.
\newblock Revise and resubmit: How real-time parsing limitations influence
  grammar acquisition.
\newblock \emph{Cognitive Psychology}, 80:73--108.

\bibitem[{Rajagopal et~al.(2022)Rajagopal, Zhang, Gamon, Jauhar, Yang, and
  Hovy}]{rajagopal-etal-2022-one}
Dheeraj Rajagopal, Xuchao Zhang, Michael Gamon, Sujay~Kumar Jauhar, Diyi Yang,
  and Eduard Hovy. 2022.
\newblock \href {https://aclanthology.org/2022.lrec-1.591} {One document, many
  revisions: A dataset for classification and description of edit intents}.
\newblock In \emph{Proceedings of the Thirteenth Language Resources and
  Evaluation Conference}, pages 5517--5524, Marseille, France. European
  Language Resources Association.

\bibitem[{Rawat and Barres(2022)}]{rawat2022real}
Mrinal Rawat and Victor Barres. 2022.
\newblock \href {https://arxiv.org/abs/2208.06802} {Real-time caller intent
  detection in human-human customer support spoken conversations}.
\newblock In \emph{Communication in Human-AI Interaction Workshop}.

\bibitem[{Robin(1996)}]{robin-1996-evaluating}
Jacques Robin. 1996.
\newblock \href {https://doi.org/10.3115/981863.981891} {Evaluating the
  portability of revision rules for incremental summary generation}.
\newblock In \emph{34th Annual Meeting of the Association for Computational
  Linguistics}, pages 205--214, Santa Cruz, California, USA. Association for
  Computational Linguistics.

\bibitem[{Rohanian and Hough(2021)}]{rohanian-hough-2021-best}
Morteza Rohanian and Julian Hough. 2021.
\newblock \href {https://doi.org/10.18653/v1/2021.acl-long.286} {Best of both
  worlds: Making high accuracy non-incremental transformer-based disfluency
  detection incremental}.
\newblock In \emph{Proceedings of the 59th Annual Meeting of the Association
  for Computational Linguistics and the 11th International Joint Conference on
  Natural Language Processing (Volume 1: Long Papers)}, pages 3693--3703,
  Online. Association for Computational Linguistics.

\bibitem[{Schlangen et~al.(2009)Schlangen, Baumann, and
  Atterer}]{schlangen-etal-2009-incremental}
David Schlangen, Timo Baumann, and Michaela Atterer. 2009.
\newblock \href {https://aclanthology.org/W09-3905} {Incremental reference
  resolution: The task, metrics for evaluation, and a {B}ayesian filtering
  model that is sensitive to disfluencies}.
\newblock In \emph{Proceedings of the {SIGDIAL} 2009 Conference}, pages 30--37,
  London, UK. Association for Computational Linguistics.

\bibitem[{Schlangen and Skantze(2011)}]{iu-restart}
David Schlangen and Gabriel Skantze. 2011.
\newblock \href {https://doi.org/10.5087/dad.2011.105} {{A General, Abstract
  Model of Incremental Dialogue Processing}}.
\newblock \emph{Dialogue and Discourse}, 2(1):83--111.

\bibitem[{Selfridge et~al.(2011)Selfridge, Arizmendi, Heeman, and
  Williams}]{selfridge-etal-2011-stability}
Ethan Selfridge, Iker Arizmendi, Peter Heeman, and Jason Williams. 2011.
\newblock \href {https://aclanthology.org/W11-2014} {Stability and accuracy in
  incremental speech recognition}.
\newblock In \emph{Proceedings of the {SIGDIAL} 2011 Conference}, pages
  110--119, Portland, Oregon. Association for Computational Linguistics.

\bibitem[{Sen et~al.(2023)Sen, Sennrich, Zhang, and
  Haddow}]{sen-etal-2023-self}
Sukanta Sen, Rico Sennrich, Biao Zhang, and Barry Haddow. 2023.
\newblock \href {https://aclanthology.org/2023.eacl-main.270} {Self-training
  reduces flicker in retranslation-based simultaneous translation}.
\newblock In \emph{Proceedings of the 17th Conference of the European Chapter
  of the Association for Computational Linguistics}, pages 3734--3744,
  Dubrovnik, Croatia. Association for Computational Linguistics.

\bibitem[{Silveira et~al.(2014)Silveira, Dozat, de~Marneffe, Bowman, Connor,
  Bauer, and Manning}]{silveira-etal-2014-gold}
Natalia Silveira, Timothy Dozat, Marie-Catherine de~Marneffe, Samuel Bowman,
  Miriam Connor, John Bauer, and Chris Manning. 2014.
\newblock \href
  {http://www.lrec-conf.org/proceedings/lrec2014/pdf/1089_Paper.pdf} {A gold
  standard dependency corpus for {E}nglish}.
\newblock In \emph{Proceedings of the Ninth International Conference on
  Language Resources and Evaluation ({LREC}'14)}, pages 2897--2904, Reykjavik,
  Iceland. European Language Resources Association (ELRA).

\bibitem[{Tjong Kim~Sang and
  De~Meulder(2003)}]{tjong-kim-sang-de-meulder-2003-introduction}
Erik~F. Tjong Kim~Sang and Fien De~Meulder. 2003.
\newblock \href {https://aclanthology.org/W03-0419} {Introduction to the
  {C}o{NLL}-2003 shared task: Language-independent named entity recognition}.
\newblock In \emph{Proceedings of the Seventh Conference on Natural Language
  Learning at {HLT}-{NAACL} 2003}, pages 142--147.

\bibitem[{Ulmer et~al.(2019)Ulmer, Hupkes, and
  Bruni}]{ulmer-etal-2019-assessing}
Dennis Ulmer, Dieuwke Hupkes, and Elia Bruni. 2019.
\newblock \href {https://doi.org/10.18653/v1/W19-4324} {Assessing
  incrementality in sequence-to-sequence models}.
\newblock In \emph{Proceedings of the 4th Workshop on Representation Learning
  for NLP (RepL4NLP-2019)}, pages 209--217, Florence, Italy. Association for
  Computational Linguistics.

\bibitem[{Vaswani et~al.(2017)Vaswani, Shazeer, Parmar, Uszkoreit, Jones,
  Gomez, Kaiser, and Polosukhin}]{vaswani2017attention}
Ashish Vaswani, Noam Shazeer, Niki Parmar, Jakob Uszkoreit, Llion Jones,
  Aidan~N Gomez, {\L}ukasz Kaiser, and Illia Polosukhin. 2017.
\newblock Attention is all you need.
\newblock \emph{Advances in neural information processing systems}, 30.

\bibitem[{Vaughan and McDonald(1986)}]{vaughan-mcdonald-1986-model}
Marie~M. Vaughan and David~D. McDonald. 1986.
\newblock \href {https://doi.org/10.3115/981131.981146} {A model of revision in
  natural language generation}.
\newblock In \emph{24th Annual Meeting of the Association for Computational
  Linguistics}, pages 90--96, New York, New York, USA. Association for
  Computational Linguistics.

\bibitem[{Zhang et~al.(2020)Zhang, Zhang, He, Wu, and
  Wang}]{zhang-etal-2020-learning-adaptive}
Ruiqing Zhang, Chuanqiang Zhang, Zhongjun He, Hua Wu, and Haifeng Wang. 2020.
\newblock \href {https://doi.org/10.18653/v1/2020.emnlp-main.178} {Learning
  adaptive segmentation policy for simultaneous translation}.
\newblock In \emph{Proceedings of the 2020 Conference on Empirical Methods in
  Natural Language Processing (EMNLP)}, pages 2280--2289, Online. Association
  for Computational Linguistics.

\bibitem[{Zheng et~al.(2020{\natexlab{a}})Zheng, Liu, Zheng, Ma, Liu, and
  Huang}]{zheng-etal-2020-simultaneous}
Baigong Zheng, Kaibo Liu, Renjie Zheng, Mingbo Ma, Hairong Liu, and Liang
  Huang. 2020{\natexlab{a}}.
\newblock \href {https://doi.org/10.18653/v1/2020.acl-main.254} {Simultaneous
  translation policies: From fixed to adaptive}.
\newblock In \emph{Proceedings of the 58th Annual Meeting of the Association
  for Computational Linguistics}, pages 2847--2853, Online. Association for
  Computational Linguistics.

\bibitem[{Zheng et~al.(2020{\natexlab{b}})Zheng, Ma, Zheng, Liu, and
  Huang}]{zheng-etal-2020-opportunistic}
Renjie Zheng, Mingbo Ma, Baigong Zheng, Kaibo Liu, and Liang Huang.
  2020{\natexlab{b}}.
\newblock \href {https://doi.org/10.18653/v1/2020.acl-main.42} {Opportunistic
  decoding with timely correction for simultaneous translation}.
\newblock In \emph{Proceedings of the 58th Annual Meeting of the Association
  for Computational Linguistics}, pages 437--442, Online. Association for
  Computational Linguistics.

\bibitem[{Zilberstein(1996)}]{zilberstein1996using}
Shlomo Zilberstein. 1996.
\newblock Using anytime algorithms in intelligent systems.
\newblock \emph{AI magazine}, 17(3):73--73.

\end{thebibliography}

\end{document}